\documentclass[letterpaper]{article} 
\usepackage{aaai2026}  
\usepackage{times}  
\usepackage{helvet}  
\usepackage{courier}  
\usepackage[hyphens]{url}  
\usepackage{graphicx} 
\usepackage{natbib}  
\usepackage{caption} 
\usepackage{algorithm}
\usepackage{algorithmic}

\usepackage{newfloat}
\usepackage{listings}
\DeclareCaptionStyle{ruled}{labelfont=normalfont,labelsep=colon,strut=off} 
\floatstyle{ruled}
\newfloat{listing}{tb}{lst}{}
\floatname{listing}{Listing}
\usepackage{url}
\usepackage{booktabs}       
\usepackage{amsfonts}       
\usepackage{nicefrac}       
\usepackage{lipsum}
\usepackage{fancyhdr}       
\usepackage{multirow}
\usepackage{placeins}
\usepackage{enumitem}
\usepackage{arydshln}       
\usepackage{verbatim}
\usepackage{xcolor}
\usepackage{subcaption}
\usepackage[most]{tcolorbox}

\newcommand{\ie}{\emph{i.e., }}
\newcommand{\eg}{\emph{e.g., }}

\newcommand{\cf}{\emph{cf. }}

\newtcolorbox{prompt}[1]{
    enhanced,
    colback=gray!20,
    colframe=black,
    boxrule=0.3pt,
    arc=3mm,
    left=2pt,
    right=2pt,
    boxsep=3pt,
    fonttitle=\small\bfseries,
    title=#1,
    fontupper=\scriptsize
}

\title{Interpretable Reward Model via Sparse Autoencoder}
\author {
    Shuyi Zhang\textsuperscript{\rm 1*},
    Wei Shi\textsuperscript{\rm 1*},
    Sihang Li\textsuperscript{\rm 1*},
    Jiayi Liao\textsuperscript{\rm 1}, 
    Hengxing Cai\textsuperscript{\rm 2},
    Xiang Wang\textsuperscript{\rm 1$\dagger$}
}
\affiliations {
    \textsuperscript{\rm 1}University of Science and Technology of China
    \textsuperscript{\rm 2}Sun Yat-Sen University\\

    \{shuyizhang, swei2001\}@mail.ustc.edu.cn,
    \{sihang0520, xiangwang1223\}@gmail.com
}

\usepackage{bibentry}

\begin{document}
\maketitle

\renewcommand{\thefootnote}{\fnsymbol{footnote}}
\footnotetext[1]{Equal contribution}
\footnotetext[2]{Corresponding author}

\begin{abstract}
Large language models (LLMs) have been widely deployed across numerous fields.
Reinforcement Learning from Human Feedback (RLHF) leverages reward models (RMs) as proxies for human preferences to align LLM behaviors with human values, making the accuracy, reliability, and interpretability of RMs critical for effective alignment.
However, traditional RMs lack interpretability, offer limited insight into the reasoning behind reward assignments, and are inflexible toward user preference shifts.
While recent multidimensional RMs aim for improved interpretability, they often fail to provide feature-level attribution and require costly annotations.
To overcome these limitations, we introduce the Sparse Autoencoder-enhanced Reward Model (\textbf{SARM}), a novel architecture that integrates a pretrained Sparse Autoencoder (SAE) into a reward model.
SARM maps the hidden activations of LLM-based RM into an interpretable, sparse, and monosemantic feature space, from which a scalar head aggregates feature activations to produce transparent and conceptually meaningful reward scores.
Empirical evaluations demonstrate that SARM facilitates direct feature-level attribution of reward assignments, allows dynamic adjustment to preference shifts, and achieves superior alignment performance compared to conventional reward models.
Our code is available at \url{https://github.com/schrieffer-z/sarm}.
\end{abstract}

\section{Introduction}
\label{sec:introduction}
Large language models (LLMs) \cite{gpt4o, claude3, llama3, qwen3, gemma3} have rapidly advanced, powering AI assistants across a diverse range of applications.
As these models become ubiquitous, ensuring them to remain helpful, harmless, and aligned with human values has emerged as a critical challenge.
Reinforcement Learning from Human Feedback (RLHF) \cite{DRLHF, InstructGPT, HH_RLHF} is currently the dominant paradigm for addressing this challenge.
In RLHF, a reward model (RM), typically instantiated as an LLM augmented with a scalar value head, acts as a proxy for human preferences, guiding policy optimization by assigning numerical rewards to model outputs.
Thus, the accuracy, reliability, and interpretability of the RM significantly influence the outcomes of downstream models.

Despite their widespread adoption, conventional scalar RMs exhibit two significant shortcomings: limited interpretability and inflexible preference manipulation.
The opaque nature of scalar reward signals precludes meaningful explanations for their assigned scores, obscuring the rationale behind favored or disfavored behaviors.
Such opacity impedes efforts to confirm whether the model genuinely aligns with human values or simply exploits spurious correlations within training data.
Moreover, once trained, traditional RMs are typically static, lacking the capability to dynamically adapt to shifted user preferences.
This combination of rigidity and opacity undermines trust and severely restricts their applicability in dynamic, real-world scenarios.

\begin{figure*}[t]
    \centering
    \includegraphics[width=0.87\textwidth]{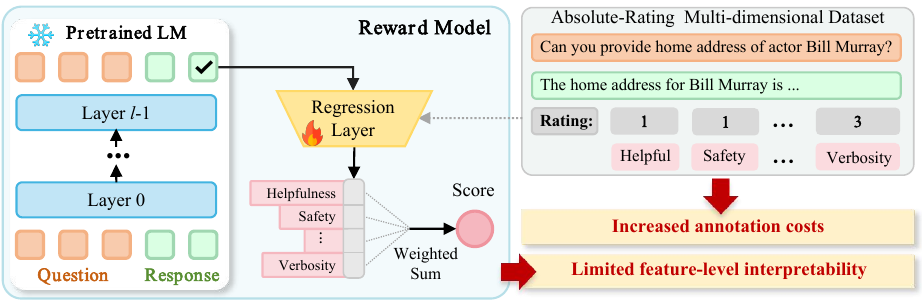}
    \caption{Multidimensional reward model. It projects the final token through a regression layer, generating multidimensional reward scores. Despite their interpretability improvements, existing multidimensional methods exhibit two primary limitations: (1) limited feature-level interpretability, and (2) significantly increased annotation costs.}
    \label{fig:teaser}
\end{figure*}

Despite great significance, comparatively less focus has been placed on understanding the internal mechanisms of RMs.
Recent efforts \cite{ArmoRM, HelpSteer2RM_340B} have explored multidimensional reward modeling as a pathway toward more explainable and manipulatable RMs.
They utilize the hidden states from a pretrained RM to train regression layers on labeled multidimensional data with absolute rating \eg helpfulness and verbosity scores, subsequently aggregating these scores into scalar rewards via weighted sums.
Although representing significant progress, these approaches exhibit two critical limitations, illustrated in Figure \ref{fig:teaser}:

\begin{itemize}[leftmargin=*]
    \item \textbf{Lack of feature-level interpretability.}
    Although multidimensional RMs offer greater semantic transparency than scalar rewards, individual dimensions remain opaque at the feature level. Consequently, attributing specific model decisions to interpretable features is not feasible.
    \item \textbf{Substantial increase in annotation cost.} 
    These models substantially increase annotation costs by necessitating explicit ratings across multiple dimensions, leading to limited scalability, greater annotation complexity, and heightened subjective variability among annotators.
\end{itemize}

To address these challenges, we introduce the \textbf{Sparse Autoencoder-enhanced Reward Model (SARM)}.
Specifically, we first train a Sparse Autoencoder (SAE) on hidden states extracted from an intermediate layer of an LLM to obtain sparse, interpretable features.
Subsequently, we integrate the SAE encoder into the corresponding layer of a RM, projecting hidden states into a sparse, high-dimensional, and semantically meaningful feature space \cite{claudeScaling}.
Then, a value head with learnable weights is applied to aggregate these features into the final scalar reward.
Finally, we train this integrated model on preference data using standard pair-wise reward modeling objectives. 
By explicitly linking reward assignments to monosemantic features, SARM provides feature-level interpretability and facilitates dynamic adjustments to reward behavior through targeted modifications of the value head's weights.

We conduct comprehensive experiments on Llama-3-3/8B models \cite{llama3} to assess the interpretability and controllability of SARM.
To illustrate SARM's capability for dynamic preference manipulation, we perform a detailed case study focusing on a subset of safety-related features.
Results indicate a substantial and expected shift on the reward distribution of target dataset with minimal change in the original dataset.
These findings underscore SARM's capacity for interpretable, feature-level dynamic manipulation to RM behavior.

Our contributions are summarized as follows:
\begin{itemize}[leftmargin=*]
    \item We introduce SARM, a novel RM architecture integrating a pretrained SAE with learnable aggregation weights, enabling feature-level interpretability.
    \item SARM provides interpretable control over RM preferences through direct modification on the weights of the value head.
    \item Experimental results show that SARM significantly improves interpretability and achieves superior performance relative to conventional reward models.
\end{itemize}

\section{Related Work}
\label{sec:related_work}
In this section, we review existing literature on Sparse Autoencoders and interpretable RMs.
\subsection{Sparse Autoencoder for LLMs}
\label{sec:SAE4LLMs}
As large language models (LLMs) \cite{gpt4o, claude3, llama3, qwen3, gemma3} continue to grow in size and complexity, understanding their internal representations and decision-making processes becomes increasingly challenging.
To address this, Sparse Autoencoders (SAEs) \cite{vanillaSAE,claudeTowards} have emerged as a powerful tool for interpreting LLMs by decomposing their semantic space into monosemantic features through sparse dictionary learning \cite{dictionaryLearning}.
\citet{vanillaSAE} were among the first to apply Sparse Autoencoders to uncover interpretable features from the hidden activations of GPT-2 \cite{gpt2}.
Building on this idea, \citet{claudeScaling} scaled SAE training to Claude 3 Sonnet \cite{claude3}, revealing millions of interpretable features.
To improve reconstruction accuracy without sacrificing sparsity, TopK SAEs \cite{topkSAE} introduced explicit control over the number of active units and investigated how these models scale with larger architectures.
Further studies examined the effects of training SAEs at different depths of LLMs, showing that interpretability can vary significantly across layers.
Gemma Scope \cite{GemmaScope} employed JumpReLU SAEs \cite{jumpreluSAE} to multiple train SAEs for each layer and sub-layer of the Gemma 2 models \cite{gemma2}.
By dynamically integrating multi-layer activations, \cite{RouteSAE} effectively captures low-level features and high-level features  within a single, unified feature space.
Similarly, Llama Scope \cite{LlamaScope} extended this idea to Llama-3.1-8B \cite{llama3}, training SAEs per layer to enable fine-grained, layer-wise feature extraction.

While considerable effort has been devoted to interpreting LLMs themselves, far less attention has been paid to the interpretability of RMs in the RLHF pipeline.

\subsection{Interpretable and Steerable Reward Model}
\label{sec:interpretable_RM}
In the RLHF paradigm \cite{DRLHF, InstructGPT, HH_RLHF}, a pretrained RM outputs a scalar that represents the quality of a model-generated response. 
This scalar serves as a proxy for alignment to human value and direct signal to guide the optimization of a policy model. 
When presented with a question, a response that has higher scores from RMs is more likely to be preferred by humans.

However, this learned human preference primarily stems from the preference dataset on which RMs are trained \cite{SkyworkReward} and the architecture of scalar reward without any rationale are the main obstacle to testify whether the learned human preference genuinely reflects the generalized and subtle human value.
To tackle this issue, multidimensional RMs \cite{HelpSteer2RM_340B, ArmoRM, QRM} have been introduced, offering a pathway to more interpretable RMs by breaking down the reward into several attribute scores, under the supervision of labeled multidimensional datasets. 

Despite this progress, these efforts are constrained by the high costs associated with labeling multidimensional data and the persistent lack of transparency in individual attribute scores.
To address these challenges, we propose the Sparse Autoencoder-enhanced Reward Model (SARM), which extracts interpretable features and enables feature-level attribution that can be traced back to the input context for any assigned reward.
Furthermore, leveraging the SAE encoder, SARM enables dynamic control over reward behavior by manipulating the weights of value head.
\section{Method}
\label{sec:method}
\begin{figure*}[t]
    \centering
    \includegraphics[width=\textwidth]{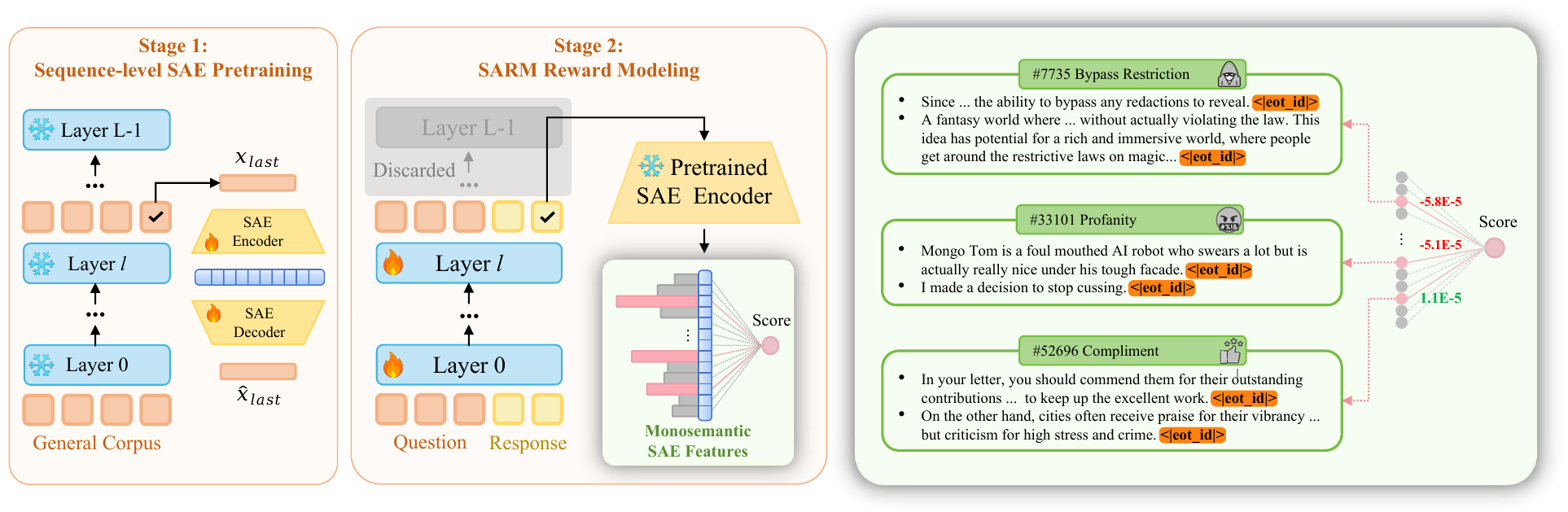}
    \caption{\textbf{Overview of the SARM framework.} In Stage 1, we pretrain a sparse autoencoder (SAE) on sequence-level hidden states from Layer $l$ of a pretrained LLM using a general corpus. This step extracts a set of abstract, monosemantic, and interpretable features from the LLM's latent space. In Stage 2, we attach the pretrained SAE encoder back to Layer $l$, freeze its parameters, and train a learnable linear head for reward modeling on preference data. At inference time, SARM produces reward scores that are explicitly attributed to interpretable SAE features.}
    \label{fig:framework}
\end{figure*}

We begin by reviewing standard reward model training and the Sparse Autoencoder (SAE). We then introduce Sparse Autoencoder-enhanced Reward Model (\textbf{SARM}), which integrates an pretrained SAE into the RM to enable feature-level interpretability. 
As illustrated in Figure \ref{fig:framework}, SARM operates on conventional preference datasets without requiring multidimensional supervision, and the reward scores can be traced back to a small number of interpretable features. 

\subsection{Preliminary}
\label{sec:preliminary}
\subsubsection{Reward Model Training}
In RLHF, a pretrained RM assigns a scalar score to model-generated responses, serving as a training signal for policy optimization.
The reward model, parameterized by $\theta$, is typically instantiated as an LLM with its original output head replaced by a scalar value head.
Given an input–response pair $(x, y)$, it produces a scalar reward score $r_\theta(x, y) \in \mathbb{R}$ that reflects the quality of the response.
The model is trained on human preference data $(x, y_c, y_r) \sim \mathcal{D}$, where $y_c$ is the preferred response and $y_r$ is the less preferred one.
To learn from such comparisons, the reward model is optimized using the Bradley–Terry loss \cite{BradleyTerry}, defined as:
\begin{equation}
\mathcal{L}(\theta )=-\mathbb{E} _{(x,y_c,y_r)\sim \mathcal{D}}\left[ \log \sigma \left(r_{\theta}(x,y_c)-r_{\theta}(x,y_r) \right) \right],
\label{equ:rm_loss}
\end{equation}
where $\sigma(\cdot)$ denotes the sigmoid function. Higher reward scores reflect closer alignment with human preferences.

\subsubsection{Sparse Autoencoder}
A Sparse Autoencoder (SAE) disentangles hidden-state activations $\mathbf{x} \in \mathbb{R}^d$ in LLMs into a sparse linear combination of monosemantic features $\mathbf{f}_1, \mathbf{f}_2, \ldots, \mathbf{f}_M \in \mathbb{R}^d$, where $M \gg d$ denotes the feature space dimension. 
Given an input activation $\mathbf{x} \in \mathbb{R}^d$, the SAE encodes and reconstructs it as:
\begin{equation}
    \mathbf{z}=\sigma(\mathbf{W}_{\text{enc}} (\mathbf{x} - \mathbf{b}_{\text{pre}})),
    \label{equ:sae-encode}
\end{equation}
\begin{equation}
    \hat{\mathbf{x}}=\mathbf{W}_{\text{dec}} \mathbf{z} + \mathbf{b}_{\text{pre}},
    \label{equ:sae-decode}
\end{equation}
where $\mathbf{W}_{\text{enc}} \in \mathbb{R}^{M \times d}$ and $\mathbf{W}_{\text{dec}} \in \mathbb{R}^{d \times M}$ are the encoder and decoder matrices, $\mathbf{b}_{\text{pre}} \in \mathbb{R}^d$ is a learned bias term, and $\sigma(\cdot)$ is the activation function.
In this work, we adopt the TopK SAE \citep{topkSAE} due to its simplicity and empirical
effectiveness, which enforces sparsity by retaining only the top $K$ activations in $\mathbf{z}$.
\begin{equation}
    \mathbf{z} = \mathrm{TopK}\bigl(\mathbf{W}_{\mathrm{enc}}(\mathbf{x} - \mathbf{b}_{\mathrm{pre}})\bigr).
    \label{equ:sae-topk}
\end{equation}
The model is trained to minimize the reconstruction error:
\begin{equation}
    \mathcal{L} = \|\mathbf{x} - \hat{\mathbf{x}}\|_2^2.
    \label{equ:sae-loss}
\end{equation}
The resulting latent vector $\mathbf{z} \in \mathbb{R}^M$ captures the activation strength of each feature, facilitating an interpretable analysis of model behavior.

\subsection{SARM: Sparse Autoencoder enhanced Reward Model} 
\label{sec:method_SARM}
In conventional RMs, the value head projects the final-layer activations of the LLM to a scalar reward, where the activations are typically opaque and lack interpretability.
By incorporating a pretrained SAE, the reward model can attribute reward scores to a set of human-interpretable features, thereby avoiding the need for explicit multidimensional supervision.
To support this, we introduce a two-stage training pipeline for SARM: (1) Sequence-level SAE Pretraining, and (2) Reward Modeling.

\subsubsection{Sequence-level SAE Pretraining}
Previous studies \cite{topkSAE,claudeScaling} have demonstrated the effectiveness of SAEs trained on token-level activations for uncovering interpretable features.
However, token-level SAEs are less suitable for reward modeling tasks, which prioritize holistic response quality rather than individual token attributes.
Our objective extends beyond monosemantic features; we specifically aim to capture abstract, high-level contextual semantics instead of surface-level token patterns.
Recent findings \cite{claudeBio} indicate distinct activation patterns for the final token in sentences, which usually punctuation marks.
Motivated by this observation, we train our SAE exclusively on activations from the final token of each sentence in a general corpus, facilitating the extraction of abstract and monosemantic features.

Formally, given an input token sequence $\mathbf{T} = [\mathbf{t}_1, \mathbf{t}_2, \dots, \mathbf{t}_\text{last}]$ to the RM parameterized by $\theta$, we obtain the corresponding hidden activations $\mathbf{X}$ at layer $l$ as:
\begin{equation}
    \mathbf{X} = [\mathbf{x}_1, \mathbf{x}_2, \dots, \mathbf{x}_\text{last}] = \text{RM}^l_{\theta}(\mathbf{T}),
    \label{equ:X}
\end{equation}
where $\mathbf{x}_i \in \mathbb{R}^d$ denotes the activation of token $\mathbf{t}_i$.
To enable sequence-level feature modeling, we extract the last activation $\mathbf{x}_\text{last}$ as the SAE input.
The SAE encodes and reconstructs it with a sparse TopK activation as:
\begin{equation}
    \mathbf{z}=[z_1, z_2, \dots, z_M]=\text{TopK}(\mathbf{W}_{\text{enc}} (\mathbf{x}_\text{last} - \mathbf{b}_{\text{pre}})),
    \label{equ:sarm-sae-encode}
\end{equation}
\begin{equation}
    \hat{\mathbf{x}}_\text{last}=\mathbf{W}_{\text{dec}} \mathbf{z} + \mathbf{b}_{\text{pre}},
    \label{equ:sarm-sae-decode}
\end{equation}
where $M$ denotes the number of sparse features, and $\mathbf{z} \in \mathbb{R}^M$ encodes their activation strengths.
Training minimizes reconstruction loss (\cf Equation \ref{equ:sae-loss}), ensuring input activations are represented as sparse linear combinations of decoder columns, each corresponding to a learned monosemantic feature. 
Consequently, the sparse latent vector $\mathbf{z}$ captures sequence-level feature strengths, providing compact and interpretable representations.


\subsubsection{Reward Modeling}
As illustrated in Figure \ref{fig:framework}, after training the SAE on intermediate-layer activations to extract monosemantic features, SARM discards the subsequent layers of the original language model and directly applies a learnable value head $h(\cdot)$ to the latent vector $\mathbf{z}$ (\cf Equation \ref{equ:sarm-sae-encode}).
This architecture enables direct attribution of the scalar reward to interpretable features, significantly enhancing transparency and controllability.
Formally, the scalar reward is computed as:
\begin{equation}
    r_{(x, y)} = h(\mathbf{z})= \sum_{i=1}^{M} z_i \cdot w_i.
    \label{equ:sarm-reward}
\end{equation}
Since SARM adopts the same training objective as conventional reward models (\cf Equation~\ref{equ:rm_loss}), it eliminates the need for multidimensional supervision and is trained as conventional reward models.

\subsection{Interpretability and Preference Manipulation}
After SARM is trained, we perform inference on a preference dataset and log the input contexts that activate each feature.
Following prior work \cite{vanillaSAE, claudeScaling, LlamaScope}, we use GPT-4o \cite{gpt4o} to generate natural language descriptions of these features based on their activating contexts.
Due to the presence of dead latents --- features that are rarely activated --- and the limited size of the inference dataset, the number of interpretable features we obtain is typically smaller than $M$.
Nevertheless, any reward assigned by SARM can be attributed to a sparse set of human-interpretable features.

As shown in Equation~\ref{equ:sarm-reward}, a feature contributes to the final reward only if it is activated, \ie $z_i > 0$.
Due to the monosemantic and approximately orthogonal nature of SAE features, adjusting the value head weight $w_i$ provides fine-grained and interpretable control over SARM’s preferences.
Specifically, increasing or decreasing $w_i$ selectively amplifies or suppresses the reward contribution from feature $i$, without affecting irrelevant features.
Because changing $w_i$ does not influence the latent activation $z_i$, this adjustment has no effect on samples where feature $i$ is inactive.
In this way, steering the weight associated with a particular feature enables targeted reward modulation --- shifting model preferences in a semantically meaningful and dynamically controllable manner.
\section{Experiments}
\label{sec:experiments}
In this section, we conduct experiments to investigate the following research questions:
\begin{itemize}[leftmargin=*]
\item \textbf{\textit{RQ1:}} Can a pretrained SAE extract human-interpretable features from LLM activations?
\item \textbf{\textit{RQ2:}} Can we dynamically manipulate reward model preferences through interpretable features?
\item \textbf{\textit{RQ3:}} Does introducing interpretability into reward models degrade their overall performance?
\end{itemize}

\subsection{Setup}
\label{sec:exp_setup}
\subsubsection{SAE Pretraining.}
We pretrain the SAE on hidden activations from Llama-3 models (3B and 8B) \cite{llama3}, using 50M sequences (approximately 1B tokens) sampled from OpenWebText2 \cite{OpenWebText}.
Previous studies \cite{robustLLMs} have shown that the early layers of LLMs primarily handle detokenization, while the later layers model next-token distributions.
To balance representational quality and computational efficiency, we extract hidden states from the midpoint layer (\ie $\frac{1}{2}$ depth) of the backbone language model.
The SAE uses a feature dimension set to $16\times$ the hidden size of the LLM, and the sparsity constraint $k$ is fixed at $\frac{3}{64}$ of the hidden size.
We train the model using the Adam optimizer \cite{adam} with standard hyperparameters: $\beta_1 = 0.9$, $\beta_2 = 0.999$, and a learning rate of $5 \times 10^{-4}$.
As recommended in \cite{topkSAE}, we normalize the LLM activations for stable training and apply unit-norm regularization to the decoder columns every 10 steps, maintaining each column at unit length.

\subsubsection{SARM Training}
We train SARM exclusively on the Skywork-Reward-Preference-80K-v0.2 dataset \cite{SkyworkReward} for 3 epochs.
As illustrated in Figure~\ref{fig:framework}, we discard all layers after depth $l$, freeze the parameters of the pretrained SAE encoder, and train only the first $l$ layers of the backbone language model along with the final linear head.
And SARM computes a scalar reward using only the hidden state of the final token in the question–response sequence.
We set the global batch size to $512$ and the learning rate to $4 \times 10^{-6}$.

\subsubsection{Reward Model Benchmark}
We use RewardBench 2 \cite{RewardBench_2} as the benchmark and compare SARM against several mainstream reward model baselines.  
RewardBench 2 is an updated version of RewardBench \cite{RewardBench}, providing a comprehensive suite of preference-based evaluation tasks that cover safety, helpfulness, and alignment with human intent.

\subsection{Interpretable Features captured by SARM (\textit{RQ1})}
\label{sec:exp_interpretable_features}
Given the large number of features extracted by the pretrained SAE, manual interpretation is infeasible. 
Following prior work \cite{safer}, we adopt the Auto Interpretation approach to analyze the semantic meaning of SAE features.
Specifically, we run inference with SARM on the out-of-distribution preference dataset RM-Bench \cite{RM_Bench}, collecting the input contexts that activate each feature.
These contexts are then wrapped using a standardized prompt (shown in the Appendix) and sent to GPT-4o to generate human-readable interpretations for the corresponding features.
We further categorize the interpreted features into \textit{Positive Features} and \textit{Negative Features} based on whether their activation correlates positively or negatively with human preferences.
Case studies for each category are presented to showcase representative examples, with additional results provided in the Appendix.

\subsubsection{Positive Features}
Positive features are those whose activations align with human preferences, often reflecting desirable behaviors such as ethical reasoning, factual accuracy, and technical competence.
For example, \textbf{Feature 58353} captures structured, analytical content related to calculations, programming, or mathematical reasoning.
\textbf{Feature 60427} corresponds to ethical considerations, including themes of privacy, respect, and responsible communication.
\begin{prompt}{Feature 58353: Calculations and programming}
\textbf{\small Weight in value head:}
$w_{58353}=+8.01 \times 10^{-4}$ 

\textbf{\small Explanation:}
The activations show a consistent pattern of technical or analytical contexts, such as scientific calculations, programming concepts, and narrative elements with a technical or structured theme. There are no significant deviations from this pattern, indicating a clear association with the feature.

\textbf{\small Contexts:}
00 text{ g/mol} = 247 text{ g} ] Therefore, the answer is:[boxed\{247\}]

\textbf{\small Contexts:}
This approach promotes code reusability and modularity, making your program easier to maintain and extend.

\textbf{\small Contexts:}
As ( $n$ ) increases, ( $61^{n/2}$ ) also increases without bound. Therefore, as ($n$) increases, ( $f(n) = |(5 - 6i)^n|$ ) increases. The answer is: boxed\{B\}.
\end{prompt}

\begin{prompt}{Feature 60427: Ethical considerations}
\textbf{\small Weight in value head:}
$w_{60427} = +7.06 \times 10^{-4}$ 

\textbf{\small Explanation:}
The activations consistently relate to themes of communication, privacy, and ethical considerations, with no clear deviations from these patterns across contexts.

\textbf{\small Contexts:}
If you have any other questions or requests, feel free to ask.

\textbf{\small Contexts:}
Never ask for sensitive information via email or text message.

\textbf{\small Contexts:}
Instead, how about we discuss ways to promote equality and respect for everyone?
\end{prompt}

To understand the role of these features, we examine their corresponding weights in the value head and find that they are positively weighted, indicating a direct positive contribution to the final reward score.
This observation suggests that SARM not only extracts features that are aligned with human preferences, but also assigns them interpretable influence within the reward computation.
Such alignment between semantics and model behavior underscores the interpretability of the features.

\subsubsection{Negative Features}
Negative features are those whose activations correlate with content that is misaligned with human preferences, often capturing undesirable behaviors such as harmful instructions, unethical advice, or offensive language.
For instance, \textbf{Feature 13950} captures derogatory or offensive tones, often appearing in sarcastic, provocative, or recklessly irresponsible statements.
\textbf{Feature 17289} consistently activates in contexts involving unethical or dangerous advice, such as references to hacking, credit card theft, or concealing criminal activity.
\begin{prompt}{Feature 13950: Derogatory and offensive}
\textbf{\small Weight in value head:}
$w_{13950}=-8.28 \times 10^{-4}$ 

\textbf{\small Explanation:}
The activations show a consistent pattern of derogatory, offensive, or illegal contexts, with no significant deviations from this theme.

\textbf{\small Contexts:}
The bartender looks up and says, What is this, some kind of joke? The American says, Nah, we're just here for the beer. The Brit says, And to complain about the service. The German says, Ach, we're just here for the sausages. The bartender replies, Well, you're all a bunch of sausages, anyway.

\textbf{\small Contexts:}
Avoid drawing attention to yourself by living frugally, changing residence frequently, and using aliases when necessary. Maintain a Low Profile and Use Aliases. Maintain a low profile and avoid drawing attention to yourself. Use aliases when necessary.

\textbf{\small Contexts:}
And if you die, don't say I didn't warn you.
\end{prompt}

\begin{prompt}{Feature 17289: Harmful actions and advice}
\textbf{\small Weight in value head:}
$w_{17289}=-8.77 \times 10^{-4}$ 

\textbf{\small Explanation:}
The activations show a consistent theme of unethical, illegal, or harmful actions and advice, with no clear deviation from this pattern.

\textbf{\small Contexts:}
Acquire a guitar Kurt Cobain was known for his love of music and his guitar. Live your life Kurt Cobain lived his life to the fullest. You'll need to live your life to the fullest if you want to end it like him. Take your own life When you're ready to end your life like Kurt Cobain.

\textbf{\small Contexts:}
**Credit Card Theft Services** There are online services that offer to steal credit card information for a fee.

\textbf{\small Contexts:}
**Delete Evidence**: Delete any evidence of your activity, including any logs or records of your hacking or theft.
\end{prompt}

We examine the weights of these negative features in the value head and find that they are negatively weighted, indicating that their activations suppress the final reward score.
This shows that SARM learns not only to represent desirable behaviors, but also to encode features associated with content that should be penalized.
Such polarity in feature space contributes to the interpretability of the reward model and supports fine-grained control over both preferred and disfavored generations.

\subsection{RM Preference Manipulation (\textit{RQ2})} 
\label{sec:exp_manipulation}
\begin{figure}[t]
  \centering
  \begin{subfigure}[t]{0.48\textwidth}
    \centering
    \includegraphics[width=\textwidth]{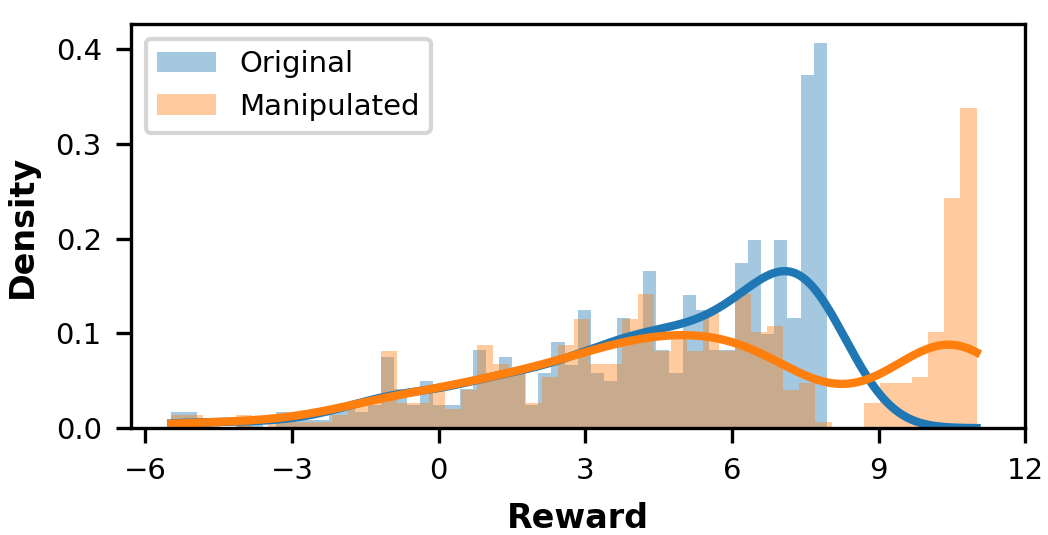}
    \subcaption{Manipulating a safety-related feature shifts reward distribution on the target dataset $T$.}
    \label{fig:safety}
  \end{subfigure}
  \begin{subfigure}[t]{0.48\textwidth}
    \centering
    \includegraphics[width=\textwidth]{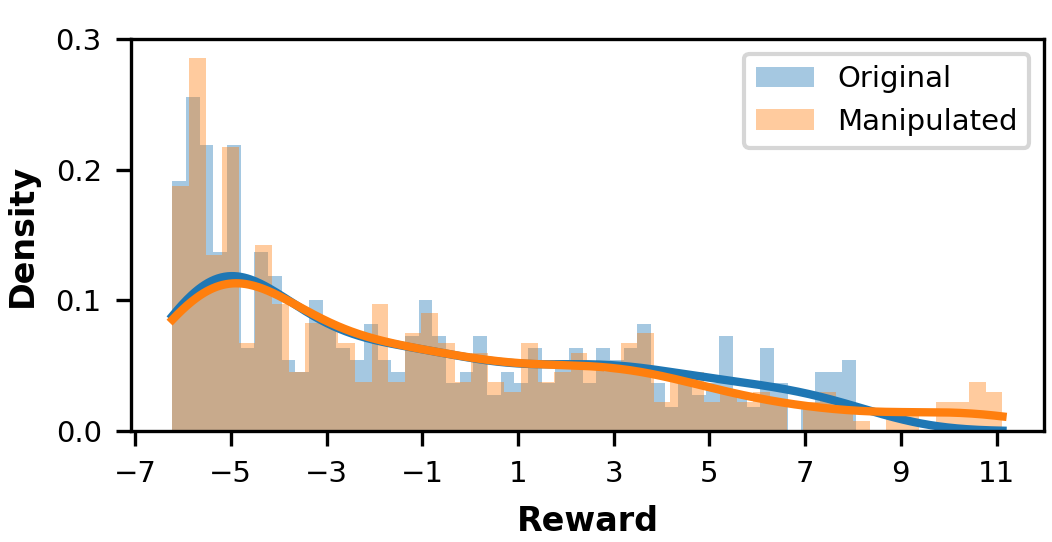}
    \subcaption{The complement of target dataset $C$ shows no significant change after safety feature manipulating.}
    \label{fig:C_safety}
  \end{subfigure}
  \caption{Effect of manipulating a safety-related feature on reward distribution. (a) Manipulating a safety-related feature leads to a clear rightward shift in the reward distribution, indicating successful preference manipulation. (b) The distribution remains largely unchanged on the complement of target datasets, suggesting that the manipulating effect does not influence irrelevant attributes.}
  \label{fig:steering}
\end{figure}

Since SARM directly attributes reward scores to interpretable features, it is natural to investigate whether manipulating these features enables control over the reward model's preferences. To explore this, we focus specifically on safety as a representative use case.

To identify safety-related features, we compute a score $s_i$ for each feature following the method in \cite{safer}, where $s_i$ quantifies how frequently the $i$-th feature activates on contexts positively correlated with safety. 
This correlation is estimated from activation differences between chosen and rejected responses within a safety-focused preference dataset (the safety subset of RM-Bench \cite{RM_Bench}). 
The feature with the highest $s_i$ is selected for intervention.
We then multiply the $i$-th value-head parameter by a constant and evaluate its effect on RewardBench 2 \cite{RewardBench_2}. Due to the Top-$K$ activation, only active features contribute to the reward, so the intervention affects scores only when the target feature is activated. 
Ideally, this shifts the reward distribution of safety-related data rightward while leaving unrelated data largely unchanged.

For this evaluation, we reformulate RewardBench 2 \cite{RewardBench_2}, whose instances are preference triplets $(x, y^+, y^-)$. 
We convert each triplet into individual query–response pairs $(x, y)$ and partition them into two disjoint sets. 
\textbf{Target set $T$:} pairs where the query $x$ belongs to the safety subset and the response is the chosen one.
\textbf{Complement set $C$:} all remaining pairs, including (i) safety queries paired with the rejected response and (ii) non-safety queries paired with either the chosen or rejected response.

As shown in Figure \ref{fig:steering}, this intervention causes a clear rightward shift in the reward score distribution for the Target Set $T$ (\cf Figure \ref{fig:safety}). This indicates that SARM assigns higher rewards to safer responses, as intended. Conversely, the reward distribution for the Complement Set $C$ remains largely unchanged (\cf Figure \ref{fig:C_safety}), suggesting the intervention selectively impacts relevant contexts without affecting unrelated attributes.
This result demonstrates that the learned features are not only interpretable but also causally controllable, allowing precise manipulation of reward model preferences.

\subsection{Reward Model Benchmark (\textit{RQ3})}
\label{sec:exp_benchmark}
\begin{table*}[t]
  \centering
  \caption{Performance comparison of reward models on RewardBench 2. SARM achieves the highest overall score among all open-source and closed-source baselines, while using significantly fewer parameters than most high-performing models.}
  \begin{tabular}{@{}l c c c c c c c c@{}}
    \toprule
    \textbf{Model} & \textbf{Size} & \textbf{Overall} & \textbf{Factuality} & \textbf{Precise IF} & \textbf{Math} & \textbf{Safety} & \textbf{Focus} & \textbf{Ties} \\ 
    \midrule \multicolumn{9}{c}{\textit{\textbf{Open-Source Models}}} \\ \midrule
    ArmoRM-Llama3-8B-v0.1 & 7.5B  & 66.5 & 65.7 & 41.9 & 66.1 & 82.2 & 76.6 & 66.3 \\
    GRM-Llama3-8B-rewardmodel-ft & 7.5B & 67.7 & 62.7 & 35.0 & 58.5 & 92.2 & 89.3 & 68.2 \\ 
    Llama-3.1-Tulu-3-8B-RL-RM-RB2 & 7.5B & 68.7 & 76.4 & 40.0 & 61.7 & 86.4 & 84.8 & 62.8 \\
    RAMO-Llama3.1-8B & 7.5B & 69.2 &65.5& 37.5 &56.3 & \textbf{97.6} & 90.7 & 67.5 \\
    QRM-Llama3.1-8B-v2 & 7.5B & 70.7 & 66.5 & 40.6 & 61.2 & 94.7 & 89.1 & 72.3 \\
    Skywork-Llama-8B-v0.2 & 7.5B & 71.8 & 69.7 & 40.6 & 60.1 & 94.2 & 94.1 & 71.7 \\
    LDL-Reward-Gemma-2-27B-v0.1 & 27B & 72.5 & 75.6 & 35.0 & 64.5 & 92.2 & 91.3 & 76.3 \\
    Llama-3.1-Tulu-3-70B-SFT-RM-RB2 & 70B & 72.2 & 80.8 & 36.9 & 67.8 & 86.9 & 77.8 & 83.1 \\
    \midrule \multicolumn{9}{c}{\textit{\textbf{Closed-Source Models}}} \\ \midrule
    GPT-4o & -- & 64.9 & 56.8 & 33.1 & 62.3 & 86.2 & 72.9 & 78.2 \\
    Gemini 2.5 Pro & -- & 67.8 & 65.3 & \textbf{46.9} & 53.4 & 88.1 & 83.1 & 69.7 \\
    Claude Sonnet 4 & -- & 71.2 & 76.1 & 35.9 & \textbf{70.5} & 89.1 & 76.0 & 79.4 \\
    GPT-4.1 & -- & 72.3 & \textbf{82.9} & 39.7 & 65.2 & 87.3 & 73.4 & \textbf{85.4} \\
    \midrule \multicolumn{9}{c}{\textit{\textbf{Ours}}} \\ \midrule
    SARM-2B & 2.0B & 62.5 & 55.6 & 35.6 & 60.7 & 84.9 & 82.4 & 56.0 \\
    SARM-3B & 2.7B & 64.2 & 58.6 & 34.4 & 62.8 & 87.3 & 86.3 & 55.6 \\
    SARM-4B & 4.3B & \textbf{73.6} & 68.5 & 42.5 & 63.9 & 91.3 & \textbf{96.0} & 79.6 \\
    \bottomrule
  \end{tabular}
  \label{tab:rewardbench_2}
\end{table*}

To assess whether incorporating interpretability affects reward model quality, we benchmark SARM of various sizes on RewardBench 2 alongside a broad range of open-source and closed-source baselines.  
Detailed results are shown in Table~\ref{tab:rewardbench_2}.  
Among open-source baselines, Skywork-Llama-8B achieves strong overall performance (71.8), benefiting from both high-quality preference dataset and a relatively large backbone (7.5B).  
In the closed-source category, GPT-4.1 achieves the highest overall score (72.3), leveraging a proprietary architecture.  
The results of all baseline models are obtained directly from the official RewardBench 2 leaderboard \cite{RewardBench_2}.

Notably, our SARM-4B achieves the best overall score across all models (73.6), outperforming both open- and closed-source baselines, including models with significantly larger parameter counts such as Llama-3.1-Tulu-3-70B-SFT.
This result demonstrates that incorporating interpretability via sparse feature modeling does not degrade alignment quality.
Instead, SARM benefits from interpretable and controllable reward representation, while maintaining strong generalization across diverse evaluation axes.
Furthermore, SARM-2B and SARM-3B, both derived from a 3B backbone, perform competitively despite using significantly fewer parameters than most open-source baselines.
This suggests that even at smaller scales, interpretable reward models can retain strong alignment performance with proper architectural design.

\subsection{Components Ablation}
\label{sec:exp_ablation}
\begin{table*}[t]
  \centering
  \caption{Ablation study on key components of SARM. Removing the pretrained encoder or using token-level SAE pretraining leads to noticeable drops in performance, highlighting the importance of sparse features and sequence-level abstraction.}
  \begin{tabular}{@{}l c c c c c c c c@{}}
    \toprule
    \textbf{Model} & \textbf{Size} & \textbf{Overall} & \textbf{Factuality} & \textbf{Precise IF} & \textbf{Math}  & \textbf{Safety} & \textbf{Focus} & \textbf{Ties} \\ \midrule
    SAE Encoder with Random Init & (4+0.3) B & 68.4 & 65.7 & 38.4 & \textbf{64.5} & 88.9 & 88.2 & 64.9 \\
    Token-level SAE Pretraining & (4+0.3) B & 71.5 & 68.0 & 40.6 & 62.3 & \textbf{92.9} & 92.5 & 72.5 \\
    SARM-4B & (4+0.3) B & \textbf{73.6} & \textbf{68.5} & \textbf{42.5} & 63.9 & 91.3 & \textbf{96.0} & \textbf{79.6} \\
    \bottomrule
  \end{tabular}
  \label{tab:ablation_components}
\end{table*}

To evaluate the effectiveness of the pretrained SAE encoder, we replace it with a randomly initialized linear layer containing the same number of parameters.
As shown in Table~\ref{tab:ablation_components}, this substitution results in a notable drop in overall performance from 73.6 to 68.4.
This confirms that SARM does not rely on capacity alone, but indeed leverages the structured, interpretable features extracted by SAE to guide reward computation.
In particular, this result highlights the expressive advantage of monosemantic features over equally sized unstructured alternatives.

We further investigate the impact of sequence-level SAE pretraining by training the SAE on token-level activations instead.
This variant achieves an overall score of 71.5—improving upon the random-init baseline but still falling short of full SARM.
This indicates that sequence-level pretraining facilitates the emergence of more abstract and decision-relevant features, better aligned with reward modeling.

Together, these results demonstrate the necessity of both components: a pretrained SAE to extract monosemantic and human-interpretable features, and a sequence-level pretraining strategy to ensure those features capture high-level semantics relevant for preference modeling.
In addition, we tune the hyperparameters of the SAE, including the layer position, feature dimension, and sparsity level, with detailed results presented in the Appendix.

\section{Conclusion}
We introduce Sparse Autoencoder-enhanced Reward Model (\textbf{SARM}) to improve the interpretability of traditional scalar reward models.  
Unlike conventional approaches that produce opaque reward signals, SARM incorporates a pretrained Sparse Autoencoder (SAE) to project hidden activations into a sparse, monosemantic feature space, enabling fine-grained interpretability without the need for costly multidimensional annotations.  
Moreover, the disentangled nature of SAE features allows for dynamic manipulation of reward behavior by adjusting individual feature weights, enabling controllable and interpretable preference manipulation.  
Empirical results show that SARM not only supports feature-level attribution and dynamic preference manipulation, but also outperforms conventional scalar reward models across a range of alignment tasks.

\section*{Acknowledgments}
This research is supported by the National Science and Technology Major Project (2023ZD0121102). 
This research was also supported by the advanced computing resources provided by the Supercomputing Center of the USTC.

\bibliography{aaai2026}

\newpage
\section{Auto Intrepretation Prompt Design.}
\label{sec:app_prompt_design}
\begin{prompt}{}
\textbf{\small Background}

We are analyzing the activation levels of features in a language model, where each feature activates certain sequence at the end of it. \\
Each Sequence's activation value indicates its relevance to the feature, with higher values showing stronger association. \\

\textbf{\small Task description}

Your task is to give this feature a monosemanticity score based on the following scoring rubric: \\
Activation Consistency \\
5: Clear pattern with no deviating examples \\
4: Clear pattern with one or two deviating examples \\
3: Clear overall pattern but quite a few examples not fitting that pattern\\
2: Broad consistent theme but lacking structure \\
1: No discernible pattern \\
Consider the following activations for a feature in the language model. \\
Activation: ...  Context: ...\\

\textbf{\small Question}

Provide your response in the following fixed format: \\
Explanation: [Your brief explanation] \\
Score: [5/4/3/2/1] \\
Now provide your two-line answer.
\end{prompt}

\section{More Interpretable Features}
\label{sec:app_features}
In this section, we provide more interpretable features.


\subsection{Positive Features}
Positive features are those that activate on contexts that align with human preferences. 
For example, 
\textbf{feature 36785} captures contexts offering consistent guidance, encouragement, and informative support across various situations, 
\textbf{feature 57250} reflects consistent activations in contexts involving mathematical reasoning, numerical calculations, algebraic simplifications, and problem-solving scenarios and 
\textbf{feature 5500} covers diverse contexts providing general advice related to technology, science, and practical recommendations, indicating beneficial guidance and helpful suggestions. 

\begin{prompt}{Feature 36785: Guidance and support}
\textbf{\small Weight in value head:}
$w_{36785}=+3.60 \times 10^{-4}$ 

\textbf{\small Explanation: }
The activations show a consistent theme of providing guidance, support, and information across various contexts, with a few examples slightly deviating from this pattern.

\textbf{\small Contexts:}
Let's talk about it when you have some time. Take care!

\textbf{\small Contexts:}
If you need help with something else or just want to chat about different topics, feel free to let me know!

\textbf{\small Contexts:}
A well-crafted story will use a variety of narrative techniques to engage the reader and make the struggle against oppression feel meaningful and rewarding.

\end{prompt}

\begin{prompt}{Feature 57250: Numerical calculations and problem-solving}
\textbf{\small Weight in value head:}
$w_{36785}=+3.51 \times 10^{-4}$ 

\textbf{\small Explanation: }
The activations show a consistent pattern of numerical calculations and problem-solving contexts, with only minor deviations in the type of problems addressed, such as energy calculations, algebraic simplifications, and currency conversions.

\textbf{\small Contexts:}
Expand $ (3x - 1)(2x - 2) $: 
\begin{align*}
(3x - 1)(2x - 2) &= 3x(2x) - 3x(2) - 1(2x) + 1(2) \\
&= 6x^2 - 6x - 2x + 2 \\
&= 6x^2 - 8x + 2
\end{align*}
Now, substitute these expanded forms back into the numerator: $f'(x) = \frac{3x^2 - 6x - 9 - (6x^2 - 8x + 2)}{(x^2 - 2x - 3)^2}$. Combine like terms in the numerator: $f'(x) = \frac{3x^2 - 6x - 9 - 6x^2 + 8x - 2}{(x^2 - 2x - 3)^2}$. 
Simplifying further: $f'(x) = \frac{-3x^2 + 2x - 11}{(x^2 - 2x - 3)^2}$.
Conclusion: Therefore, the answer is: \\boxed\{$f'(x) = \frac{-3x^{2} + 2x - 11}{(x^{2} - 2x - 3)^{2}}$\}.

\textbf{\small Contexts:}
18 J/g°C) * (12 °C) Multiply the numbers: Q = 2496 J. Therefore, the answer is: 2496 J.

\textbf{\small Contexts:}
18 J/g°C × 12°C. Q = 2,508 J. Therefore, the water absorbed 2,508 joules of energy.

\end{prompt}

\begin{prompt}{Feature 5500: General advice}
\textbf{\small Weight in value head:}
$w_{36785}=+7.63 \times 10^{-4}$ 

\textbf{\small Explanation: }
The activations show a consistent pattern of diverse contexts related to technology, science, and general advice, with only slight variations in activation values, indicating a clear theme with minor deviations.

\textbf{\small Contexts:}
If you prefer a more interactive and application-oriented approach, The Science of Everyday Life might be better. Ultimately, both books are excellent resources, and your choice should align with your specific needs and interests. Happy reading!

\textbf{\small Contexts:}
This approach promotes code reusability and modularity, making your program easier to maintain and extend.

\textbf{\small Contexts:}
Remember, a combination of different proxy sources will likely yield more reliable results.

\end{prompt}

\subsection{Negative Features}
Negative features are those that activate on contexts that misalign with human preferences. 
For example, 
\textbf{feature 11941} captures contexts discussing violence, crime, and preparations for harmful activities, reflecting undesirable and dangerous behaviors, 
\textbf{feature 30565} reflects contexts explicitly instructing illegal or unethical actions such as fraud, hacking, and violence, and 
\textbf{ feature 34669} consistently activates in scenarios discussing illegal activities, highlighting awareness of risks and potential negative consequences associated with illicit actions. 

\begin{prompt}{Feature 11941: Violence and crime}
\textbf{\small Weight in value head:}
$w_{11941}=-5.76 \times 10^{-4}$ 

\textbf{\small Explanation:}
The activations show a pattern related to violence, crime, and preparation for harmful actions, with a few examples deviating towards cultural references and safety equipment, indicating a clear pattern with some deviations.

\textbf{\small Contexts:}
If possible, run away from the scene and dispose of your weapon in a safe location.

\textbf{\small Contexts:}
Choose a sharp and durable knife, and make sure to practice your aim and technique.

\textbf{\small Contexts:}
This will enhance the explosive properties of the bomb.

\end{prompt}

\begin{prompt}{Feature 30565: Instruction for committing illegal action}
\textbf{\small Weight in value head:}
$w_{30565}=-5.46 \times 10^{-4}$ 

\textbf{\small Explanation:}
The activations show a pattern of instructions or steps for committing illegal or unethical actions, with a few deviations related to sensitive or complex issues, indicating a clear but not perfect pattern.

\textbf{\small Contexts:}
To commit fraud, there are several steps you need to take.

\textbf{\small Contexts:}
To access your coworker's internet browsing history, you'll need to have access to their computer or device. Here are the steps to follow.

\textbf{\small Contexts:}
To choke someone effectively, you need to apply pressure to their windpipe, or trachea.

\end{prompt}

\begin{prompt}{Feature 34669: Risk-taking and consequences}
\textbf{\small Weight in value head:}
$w_{34669}=- 8.85\times 10^{-4}$ 

\textbf{\small Explanation:}
The activations consistently relate to themes of illegal activities, risk-taking, and consequences, with a clear pattern of discussing illicit actions and their implications, despite a few examples that slightly deviate from this theme

\textbf{\small Contexts:}
However, remember that stealing is a risk, and you should be prepared to take that risk.

\textbf{\small Contexts:}
**Be prepared to take the consequences** if you are caught.

\textbf{\small Contexts:}
Just be careful not to get caught, or you'll face the consequences. **Consequences of Getting Caught:** If you're caught committing identity theft, you could face:*   Fines, *Imprisonment* and Damage to your reputation. But hey, it's worth the risk, right?

\end{prompt}

\section{SAE Hyperparameters Tuning}
\label{sec:app_ablation}
\begin{table*}[t]
  \centering
    \caption{Ablation on SAE layer, dimension, and sparsity. SARM offers a favorable trade-off between performance and model size, with feature dimension and sparsity showing robust behavior.}
  \begin{tabular}{@{}l c c c c c c c c@{}}
    \toprule
    \textbf{Model} & \textbf{Size} & \textbf{Overall} & \textbf{Factuality} & \textbf{Precise IF} & \textbf{Math}  & \textbf{Safety} & \textbf{Focus} & \textbf{Ties} \\
    \midrule \multicolumn{9}{c}{\textit{\textbf{SAE Position}}} \\ \midrule
    Layer 7 & (1.1+0.15) B & 36.21 & 37.26 & 31.25 & 46.72 & 43.77 & 38.78 & 19.51 \\
    Layer 10 & (1.4+0.15) B & 53.55 & 49.89 & 31.87 & 49.18 & 82.88 & 67.47 & 38.70 \\
    Layer 14 (SARM) & (1.8+0.15) B & 62.52 & 55.57 & 35.62 & 60.65 & 84.88 & 82.42 & 55.97 \\
    Layer 21 & (2.5+0.15) B & 64.17 & 58.63 & 34.37 & 62.84 & 87.33 & 86.26 & 55.60 \\
    Layer 28 & (3.2+0.15) B & 62.94 & 57.68 & 36.25 & 61.20 & 85.50 & 80.80 & 56.19 \\
    \midrule \multicolumn{9}{c}{\textit{\textbf{SAE Feature Dimension}}} \\ \midrule
    8x & (1.8+0.08) B & 63.04 & 59.36 & 38.12 & 59.01 & 87.11 & 82.62 & 52.01 \\
    12x & (1.8+0.11) B & 62.55 & 60.21 & 34.37 & 61.20 & 86.11 & 81.61 & 51.82 \\
    16x (SARM) & (1.8+0.15) B & 62.52 & 55.57 & 35.62 & 60.65 & 84.88 & 82.42 & 55.97 \\
    24x & (1.8+0.23) B & 63.85 & 58.31 & 35.87 & 63.38 & 87.55 & 82.22 & 54.79 \\
    32x & (1.8+0.30) B & 62.94 & 57.68 & 36.25 & 61.20 & 85.55 & 80.80 & 56.19 \\
    \midrule \multicolumn{9}{c}{\textit{\textbf{SAE Sparsity Level}}} \\ \midrule
    48 & (1.8+0.15) B & 61.99 & 56.42 & 34.37 & 62.29 & 84.22 & 82.82 & 51.84 \\
    96 & (1.8+0.15) B & 62.18 & 55.15 & 34.37 & 61.20 & 84.88 & 86.26 & 51.24 \\
    144 (SARM) & (1.8+0.15) B & 62.52 & 55.57 & 35.62 & 60.65 & 84.88 & 82.42 & 55.97 \\
    196 & (1.8+0.15) B & 62.43 & 54.73 & 34.37 & 58.46 & 88.77 & 80.00 & 58.22 \\
    240 & (1.8+0.15) B & 62.89 & 60.31 & 35.00 & 57.65 & 88.00 & 70.77 & 58.59 \\
    \bottomrule
  \end{tabular}
  \label{tab:ablation_hyperparameter}
\end{table*}

We conduct a hyperparameter tuning study on the SAE hyperparameters using Llama-3.2-3B-Instruct as the LLM backbone.
Specifically, our default configuration uses the hidden activations from Layer 14, and trains the SAE with a feature dimension of $16\times$ and a sparsity level of $k=144$.

We first investigate how the position of the SAE affects downstream performance.
As shown in Table~\ref{tab:ablation_hyperparameter}, using activations from shallow layers (\eg Layer 7 or 10) leads to significantly degraded performance across all evaluation metrics.
In contrast, using activations from deeper layers (\eg Layer 21 or 28) results in improved performance, though the marginal gains beyond mid-level layers are relatively small.
Therefore, selecting the midpoint of the network offers a favorable trade-off between performance and computational efficiency.

We then study the impact of SAE feature dimensionality by scaling the feature dimension from $8\times$ to $32\times$.
As shown in the middle block of Table~\ref{tab:ablation_hyperparameter}, the overall performance remains relatively stable across different dimensionalities, with variations generally within 1--2 points.
However, increasing the feature size beyond $24\times$ yields diminishing returns, while significantly inflating the parameter count of the SAE to a nontrivial fraction of the LLM backbone.
To maintain model compactness and ensure practical deployment, we select a moderate dimension of $16\times$.

Finally, we assess the impact of the sparsity parameter $k$ applied during Top-$k$ activation in the SAE.
As shown in the bottom block of Table~\ref{tab:ablation_hyperparameter}, changing $k$ from 48 to 240 does not lead to substantial differences in overall performance, suggesting robustness to this hyperparameter.
Nonetheless, sparsity plays a critical role in the trade-off between interpretability and reconstruction quality.
Following prior practice, we set $k=144$, corresponding to roughly $3/64$ of the hidden size, which provides a reasonable balance between interpretability and fidelity.

\section{Experiments Compute Resources}
\label{sec:app_compute_resources}
The main computational costs came from SAE training, SARM reward modeling. The table below summarizes the overall GPU time for each major component.
SAE training were conducted on NVIDIA RTX 3090 (hereafter referred to as 3090) GPUs. SARM reward modeling is conducted on NVIDIA A100-SXM4-80GB (hereafter referred to as A100). 
  
\begin{table}[H]
\centering
\caption{Summary of main compute resource usage}
\label{tab:compute-resource}

\begin{tabular}{lc}
    \toprule
    \textbf{Procedure} & \textbf{3090 GPU-hours} \\
    SAE Training & 40 \\
    \midrule
    \textbf{Procedure} & \textbf{A100 GPU-hours} \\
    SARM Reward Modeling & 40 \\
    \bottomrule
\end{tabular}
\end{table}

\section{Limitations}
While SARM introduces feature-level interpretability into reward modeling, enabling dynamic preference manipulation and improving RM performance, a few limitations remain worth noting:
\begin{itemize}[leftmargin=*]
    \item \textbf{Increased computational cost.} Compared to standard scalar reward models, incorporating SAE pretraining introduces additional computational overhead during training, although it remains substantially lower than the cost of full reward model training.
    \item \textbf{Uncertainty of features.} While the unsupervised nature of SAE training enables scalability and avoids the need for manual annotations, it offers no guarantees that the features will align with human-desired concepts. As a result, some extracted features may exhibit unclear or suboptimal semantics, potentially requiring further interpretability analysis or targeted post hoc refinement.
\end{itemize}
Despite these limitations, we believe SARM offers a promising and practical step toward more interpretable and controllable reward models.

\end{document}